\documentclass[letterpaper, 10 pt, conference]{ieeeconf}  

\IEEEoverridecommandlockouts                              

\usepackage{graphics} 
\usepackage{epsfig} 
\usepackage{mathptmx} 
\usepackage{times} 
\usepackage{amsmath} 
\usepackage{amssymb}  
\usepackage{hyperref}       
\usepackage{color}
\usepackage{adjustbox}
\usepackage{stfloats}
\usepackage{subcaption}

\usepackage{titlesec}
\titleformat{\paragraph}
  [runin] 
  {\normalfont\bfseries} 
  {} 
  {-0.5em} 
  {} 

\title{\LARGE \bf
{
UGotMe: An Embodied System \\
for Affective Human-Robot Interaction}
}

\author{Peizhen Li$^{1}$ Longbing Cao$^{1}$ Xiao-Ming Wu$^{2}$ Xiaohan Yu$^{1}$ Runze Yang$^{1,3}$
\thanks{$^{1}$School of Computing, Macquarie University
{\tt\small \{peizhen.li1, runze.yang\}@hdr.mq.edu.au, \tt\small \{longbing.cao, xiaohan.yu\}@mq.edu.au}
}%
\thanks{$^{2}$School of Computer Science and Engineering, Sun Yat-sen University
{\tt\small wuxm65@mail2.sysu.edu.cn}}
\thanks{$^{3}$Department of Automation, Shanghai Jiao Tong University
{\tt\small runze.y@sjtu.edu.cn}}
}

\begin{document}

\maketitle
\thispagestyle{empty}
\pagestyle{empty}


\begin{abstract}
Equipping humanoid robots with the capability to understand emotional states of human interactants and express emotions appropriately according to situations is essential for affective human-robot interaction.
However, enabling current vision-aware multimodal emotion recognition models for affective human-robot interaction in the real-world raises embodiment challenges: addressing the environmental noise issue and meeting real-time requirements. First, in multiparty conversation scenarios, the noises inherited in the visual observation of the robot, which may come from either 1) distracting objects in the scene or 2) inactive speakers appearing in the field of view of the robot, hinder the models from extracting emotional cues from vision inputs.
Secondly, real-time response, a desired feature for an interactive system, is also challenging to achieve. 
To tackle both challenges, we introduce an affective human-robot interaction system called UGotMe designed specifically for multiparty conversations. 
Two denoising strategies are proposed and incorporated into the system to solve the first issue.
Specifically, 
to filter out distracting objects in the scene, we propose extracting face images of the speakers from the raw images and introduce a customized active face extraction strategy to rule out inactive speakers.
As for the second issue, we employ efficient data transmission from the robot to the local server to improve real-time response capability. 
We deploy UGotMe on a human robot named Ameca to validate its real-time inference capabilities in practical scenarios.
Videos demonstrating real-world deployment are available at 
\href{https://lipzh5.github.io/HumanoidVLE/}{https://lipzh5.github.io/HumanoidVLE/}

\end{abstract}


\section{Introduction}

Humanoid robots and AI \cite{cao24humanoid} are increasingly engaging with people, particularly through conversational interactions, for applications like healthcare and services. To enhance the naturalness and fluidity of human-robot conversations, it is essential to endow these robots with the capability to perceive and understand human emotional states and to express emotions in a manner that resembles human behavior.
However, enabling current vision-aware multimodal emotion recognition models for affective human-robot interaction raises challenges: addressing the environmental noise issue and meeting real-time requirements. 

First, in multiparty human-robot conversation scenarios, there can be environmental noise in the visual observation of the robot.
\begin{figure}[h]
    \centering
    \includegraphics[width=0.98\linewidth]{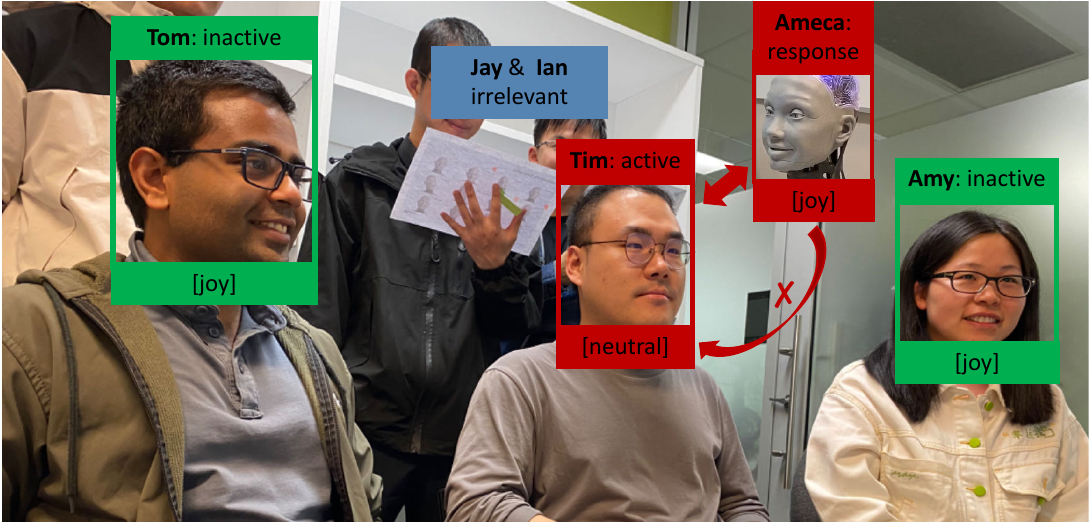}
    \caption{In a multiparty human-robot conversation scenario, the active speaker (Tim) initiates the conversation by simply stating a fact ``I am still working on my experiments" while holding a neutral face. The humanoid robot (Ameca) is supposed to deliver a neutral face in response to the speaker. However, two inactive speakers (Tom and Amy) and other persons (Jay and Ian) irrelevant to the conversation holding different facial expressions from the active speaker appear in the field of view of the robot, which may confuse the model, leading to the wrong answer.}
    \label{fig:distraction}
\end{figure}
As shown in Fig~\ref{fig:distraction}, the noises come from 1) distracting objects in the scene, like the shelf; 2) inactive speakers and other irrelevant persons who appear in the field of view of the the robot. These noises hinder the model from focusing on emotion-rich facial expressions of active speakers and extracting emotional cues from them. 
Secondly, real-time response, a desired feature for an interactive system, is also hard to achieve due to factors such as the inference speed of large deep models, communication between system modules or other hardware limitations. 

\begin{figure*}[t]
    \centering
    \includegraphics[width=0.9\linewidth]{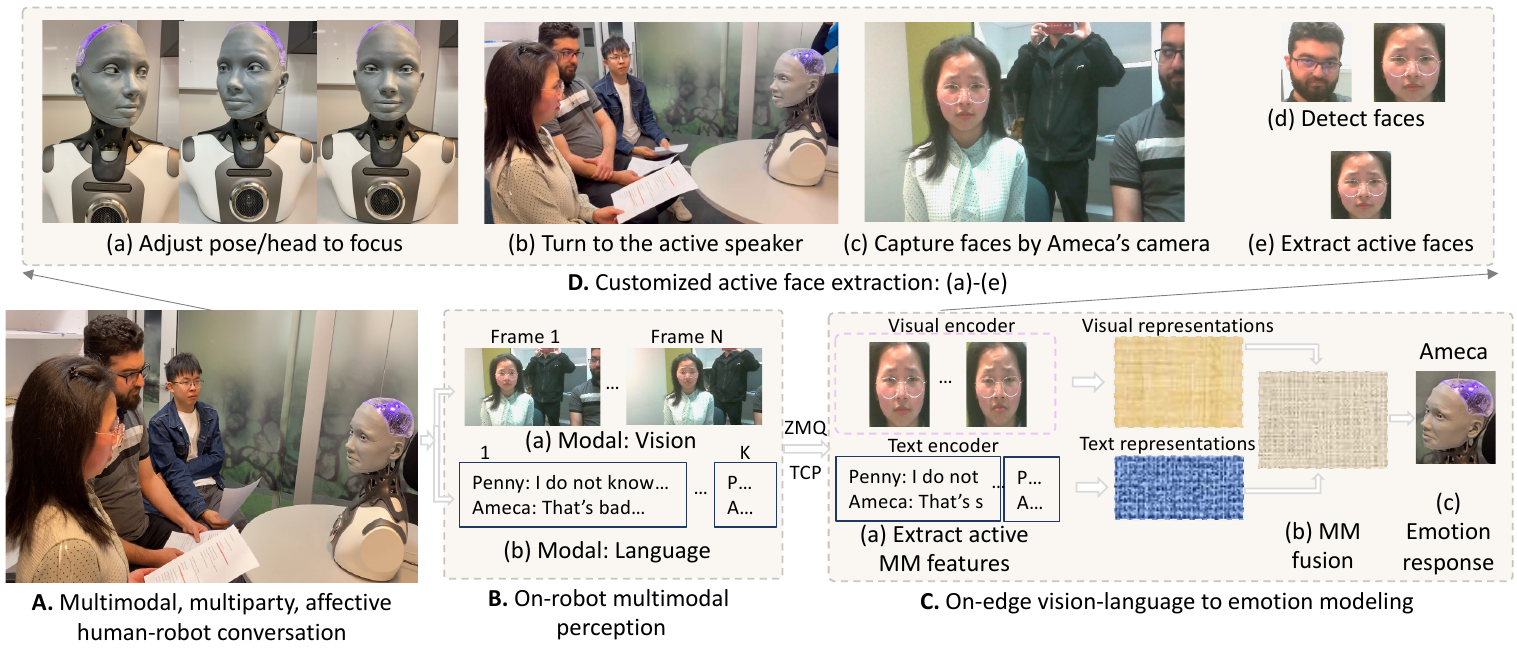}
    \caption{An overview of UGotMe, the proposed affective human-robot interaction system. The working pipeline includes on-robot multimodal perception (B) and on-edge vision-language to emotion modeling (C), where multimodal emotion recognition and robotic facial expression generation occur. D. Customized active face extraction (a)-(e) handle the environmental noise issue in 
    Fig.~\ref{fig:distraction}
     }
   
    \label{fig:ugotme}
\end{figure*}

To tackle both challenges, we introduce an affective human-robot interaction system called UGotMe, designed specifically for multiparty human-robot conversations. 
Two denoising strategies are proposed and incorporated into the system to address the issue of environmental noise.
Specifically, to filter out distracting objects in the scene, such as moving vehicles and stationary furniture, we propose extracting the faces of the speakers from raw images captured by the robot's onboard camera.
To rule out inactive speakers or persons irrelevant to the conversation, 
we employ a customized active face extraction strategy. In particular,
we adjust the robot's head pose and left-eye camera 
according to the direction of sound arrival to ensure that the active speaker is centered along the x-axis of the image. This approach allows us to accurately extract the corresponding faces.
Inspired by~\cite{baltruvsaitis2015cross}, we apply person-specific neutral normalization to the extracted face images to account for significant individual differences. 
As for the second challenge, to improve real-time response capability of the system, we stream the captured images in bytes from the robot to the local server continuously in a separate thread and buffer the latest $T$ frames, so that they can be consumed by models located on the server immediately. 
An overview of the system is shown in Fig.~\ref{fig:ugotme}, where two main stages are 
on-robot mulitmodal perception and on-edge vision-language to emotion modeling.

In particular, emotion recognition module of the system, the proposed 
Vision-Language to Emotion (VL2E) model, is designed to be compatible with the aforementioned denoising strategies, i.e., the vision encoder of VL2E consumes face sequences and extracts features at the frame level. It models intra-modal interactions within visual features using self-attention transformer.
In addition, to ensure the robot can understand human emotions in a manner similar to human interaction, we incorporate context modeling by considering the most recent $k$ utterances in the dialogue when calculating textual representations. We employ multimodal transformer to fuse features from visual and textual modalities.
On the MELD dataset, VL2E outperforms all baselines in weighted average F1.
An illustration of VL2E is shown in Fig.~\ref{fig:vle-model}. 
We aim to enable humanoid robots to convey emotions via facial expressions, as they are recognized as a highly effective means of communicating affective information compared to verbal cues and tone~\cite{mehrabian2017communication}. Therefore, upon emotion recognition, we directly map them to robotic facial expressions in line with parallel empathy (generating the same emotion as the peer)~\cite{davis2018empathy}. In the present study, we only have access to robotic facial expressions within a predefined set, corresponding to seven basic emotions (including neutral, surprise, fear, sadness, joy, disgust, and anger). These expressions can be applied to the humanoid robot named Ameca, which was developed by Engineered Arts.
We deploy UGotMe on Ameca, demonstrating its real-time inference capabilities in practical scenarios. 

Our main contributions are summarized as: 1) 
we introduce UGotMe, an affective human-robot interaction system designed to address the environmental noise issue and meet real-time requirements, thereby facilitating immediate emotional exchanges between humans and humanoid robots; 2) we propose a vision-language to emotion (VL2E) model, which can be embedded into the embodied system and also outperforms all baselines on the MELD dataset~\cite{poria2018meld};
 3) we deploy UGotMe on a physical humanoid robot named Ameca and demonstrate its practical usage in the real world.

\section{Related work}
\textbf{Emotion Recognition for Affective Human-Robot Interaction}
Research studies on emotion recognition for affective human-robot interaction can be roughly divided into two categories: vision-based and multimodal methods. Vision-based methods focus on facial expressions of human interactants and utilize them to  recognize emotions~\cite{zhu2022affective,chowdary2023deep,maeda2020human, maeda2018human, esfandbod2019human,ghorbandaei2018human,liu2017facial}.
However, these methods overlook the potential insights from other modalities, such as text and audio.
Although some multimodal emotion recognition methods exist for affective human-robot interaction~\cite{chen2022k, cid2015novel,jung2004affective}, they are not well-suited for multiparty conversation scenarios. This is because: 1) they fail to account for potential noise from the presence of multiple participants, and 2) they do not fully leverage the dialogue context, which is crucial for accurate emotion recognition in conversations~\cite{ma2020survey}. We address both limitations by introducing an affective human-robot interaction system called UGotMe.

\textbf{Vision-Aware Multimodal Emotion Recognition in Conversation}
There are many vision-aware multimodal models for emotion recognition in conversation (ERC)~\cite{yun2024telme, hu2022unimse}. By ``vision-aware", we mean that the model's input modalities include visual data.
For example, TelME considers different contributions of text and non-verbal modalities (vision and audio), and incorporates cross-modal knowledge distillation to improve the efficacy of weak non-verbal modalities.
However, these models do not account for environmental noises, particularly distracting facial expressions of inactive speakers in multiparty conversation scenarios, which impair their performance in real-world deployments.
On the contrary, some models pay attention to 
environmental noises and try to alleviate them by extracting faces of speakers~\cite{li2022emocaps, zheng2023facial, chudasama2022m2fnet}, though some of them do not consider real-time settings. For instance, FacialMMT~\cite{zheng2023facial} takes both previous and future conversational turns as inputs to compute textual representations of the current turn, which makes it infeasible for real-world deployment where future turns are not accessible.
Motivated by these approaches, we propose a Vision-Language to Emotion (VL2E) model designed to facilitate real-world affective human-robot interactions.

\section{Method}
We aim to enable current vision-aware multimodal emotion recognition models for affective human-robot interaction in the real world and introduce an interaction system called UGotMe to tackle the embodiment challenges, specifically addressing the environmental noise issue and meeting real-time requirements.
An overview of UGotMe is provided in Fig.~\ref{fig:ugotme}.
The working pipeline of the system can be divided into three phases: on-robot multimodal perception, on-edge vision-language to emotion modeling, and robotic facial expression execution.
In the on-robot multimodal perception phase, RGB images captured by the robot's onboard camera and audio signals from microphone are collected. Dialogue texts are transcribed from audio using Google Cloud speech-to-text service, which are crucial for modeling conversation context. During the on-edge vision-language to emotion modeling phase, vision and language data are transmitted to the local server, where we apply denoising strategies to filter out distracting objects and inactive speakers from raw visual perceptions. Emotional states of human interactants are then recognized based on visual and textual inputs in this phase. Robotic facial expressions are generated by 
directly mapping the recognized emotional states to the robotic facial expression within a predefined set. 
In the robotic facial expression execution phase, the humanoid robot will execute the generated expression in line with parallel empathy (generating the same emotion as the peer)~\cite{davis2018empathy}.
To facilitate real-time response, we stream the captured images in bytes continuously in a separate thread, so that they can be consumed by the model immediately.

\subsection{Denoising Strategies}
We propose two denoising strategies---face extraction and customized active face extraction---to address the embodiment issue related to environmental noise in multiparty conversation scenarios. 
Specifically, to filter out distracting objects in the scene such as moving vehicles and stationary furniture, we propose extracting the faces of the speakers from the raw RGB images captured by the robot's onboard (left-eye) camera. 
Customized active face extraction aims to extract the faces of active speakers who are conversing with the robot by leveraging its characteristics. In particular,
to exclude inactive speakers who may be present in the robot's field of view but are not speaking, we propose to adjust the robot's head pose and left-eye camera according to the direction of sound arrival so that it can focus directly on the active speaker.
We use MTCNN~\cite{zhang2016joint} for face detection. Among possible faces detected from the image, the face centered
along the x-axis will be considered as the active speaker's face and subsequently extracted for further processing.
Note that for model training and evaluation on datasets, we extract faces using OpenFace toolkit~\cite{baltruvsaitis2016openface} and all extracted faces are utilized.
Inspired by~\cite{baltruvsaitis2015cross}, we apply person-specific neutral normalization to account for significant individual differences. Specifically, given a sequence of face images for one utterance, we subtract the corresponding neutral face and feed delta images~\cite{5543148} to the vision encoder. During model training and evaluation on the MELD dataset, we manually select one neutral face for each leading role, while treating the first frame of the face sequence as the neutral face for others. In real-world deployments, where we may not always have access to the neutral faces of the speakers, we also use the first frame as the neutral face.We will try to apply or design more reasonable techniques for neural state estimation in our future work.

\subsection{Efficient Data Transmission}
Due to hardware restrictions, multimodal robotic sensory data has to be transmitted from the robot to a local server for further processing. The inevitable round-trip latency hinders the system from achieving real-time response. As a mitigation, we stream the captured RGB images in bytes to the server continuously in a separate thread, while buffering the most recent $T$ frames ($T=640$ in our implementation, corresponding to a streaming rate of 25 FPS). Textual data is transmitted only when conversational turns are generated. 
In this study, we use the ZMQ~\footnote{https://zeromq.org/} Python library to transmit both visual and textual data via TCP, enabling real-time responses.


\subsection{Emotion Recognition in Conversation}
\begin{figure}
    \centering
    \includegraphics[width=0.98\linewidth]{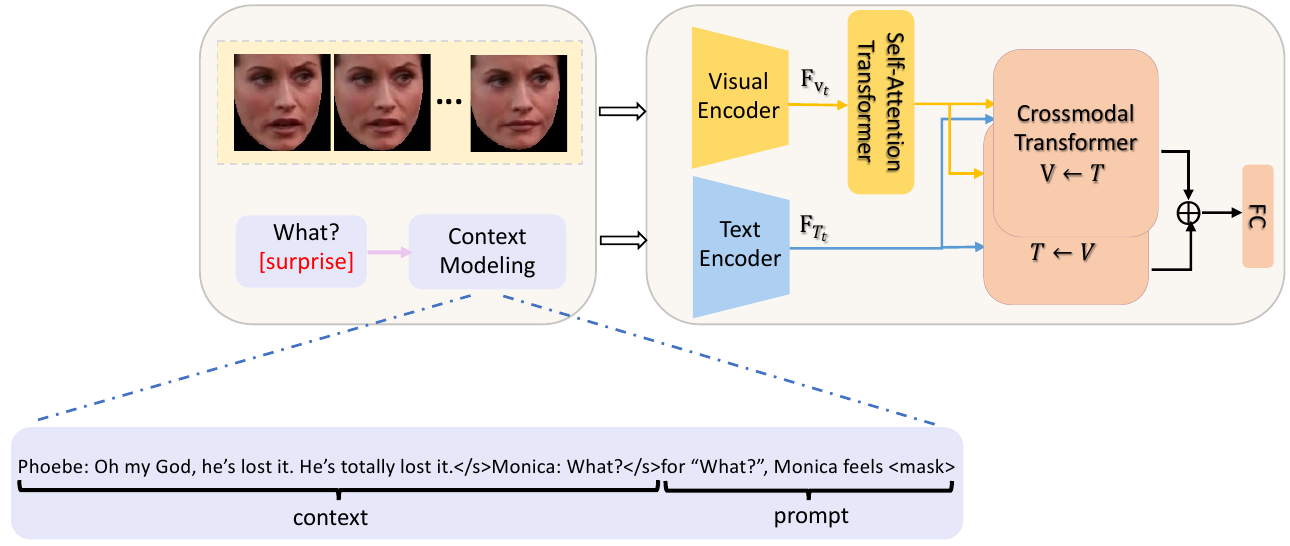}
    \caption{An illustration of the VL2E model.}
    \label{fig:vle-model}
\end{figure}

The Vision-Language to Emotion (VL2E) model dedicated to emotion recognition in multiparty conversation is designed to be  
compatible with the aforementioned denoising strategies. As shown in Fig.~\ref{fig:vle-model}, the vision encoder of the VL2E model processes face sequences of active speakers (individuals who are currently conversing with the robot) and extracts visual features at the frame level. In this way, the model can ignore the presence of irrelevant environmental noise and better focus on the emotion-rich facial expressions. 
In the present study, the InceptionResnetv1 model~\cite{szegedy2017inception} pretrained on the CASIA-WebFace dataset~\cite{yi2014learning} is utilized as the visual encoder. Extracted visual features, denoted as $F_{v_t}$, for utterance $t$ will be fed into a self-attention transformer to model intra-modal interactions.

To account for conversation context, we conduct context modeling when calculating the textual representation for the current utterance $u_t$ (where $s_t$ refers to the corresponding speaker), taking into consideration the most recent $k$ turns~\cite{song2022supervised}:
\begin{equation}
    C_t=[s_{t-k}, u_{t-k}, s_{t-k+1}, u_{t-k+1}, \cdots, s_t, u_t].
\end{equation}
To help the model distinguish between the context and the target turn, we adopt an approach inspired by prompt learning~\cite{liu2023pre}, appending a prompt $P_t$ to the conversation context for the $t$-th turn. The prompt $P_t$ is defined as follows:
\begin{equation}
    P_t = \text{for}~u_t, s_t~ \text{feels}~ \langle\text{mask} \rangle.
\end{equation}
In the present study, we employ the SimCSE~\cite{gao2021simcse} model as our text encoder, with the full input to the encoder being $C_t \oplus P_t$, where $\oplus$ refers to the concatenation operation. 
The last hidden state 
$H^k_t = \text{SimCSE}(C_t \oplus P_t)$ will be used as the context embedding, which differs from previous studies ~\cite{yun2024telme, song2022supervised}, where only the embeddings of the special token $\langle\text{mask} \rangle$ from $H_t^k$ are leveraged.
Note that we include speaker names during model training and evaluation on datasets, but not for embodied usage since speaker identities are not always accessible in real-world deployment. Therefore, conversation context and prompt in real-world deployment will be
\begin{equation*}
    \Tilde{C}_t=[u_{t-k}, u_{t-k+1}, \cdots, u_t],
\end{equation*}
and 
\begin{equation*}
    \Tilde{P}_t = \text{for}~u_t, ~\text{speaker} ~ \text{feels}~ \langle\text{mask} \rangle,
\end{equation*}
respectively.

To better model inter-modal interactions, we leverage the crossmodal transformer~\cite{tsai2019multimodal} to perform multimodal fusion once the unimodal features are extracted.


\section{Experiments}
\subsection{Setup}
In real-world deployments of UGotMe in multiparty conversation scenarios, 
a humanoid robot converses with several human participants (one person at a time). It is tasked with recognizing human emotions based on conversation context and visual observation and executing the  appropriate robotic facial expressions. 
The robot only has access to a predefined set of facial expressions that correspond to seven basic emotions: neutral, surprise, fear, sadness, joy, disgust, and anger.
We have designed a \href{https://github.com/lipzh5/AmecaVLE/blob/master/docs/misc/news\_sharing.txt}{script} for volunteers to engage in conversations with the robot. The script is used through all real-world experiments. GPT-based language models are utilized to generate textual responses for the robot. The text-to-speech service is provided by Amazon Web Services.

All experiments on the dataset are conducted on a single NVIDIA H100 GPU. We use AdamW optimizer for VL2E model training and apply \textit{cosine schedule with warmup} as the learning rate scheduler.


\subsection{Dataset}
To train and evaluate the proposed VL2E model, we use MELD~\cite{poria2018meld}, a multiparty dataset comprising over 13,000 utterances extracted from Friends TV series. It also provides emotion annotations for each utterance with seven emotion categories including neutral, surprise, fear, sadness, joy, disgust, and anger.

\subsection{Evaluation}
For the proposed VL2E and other comparison methods, we use the weighted average F1 score as the evaluation metric on the class-imbalanced dataset, MELD.

For the embodied system, UGotMe, the evaluation metrics include emotion response accuracy and user experience (rated on a scale from 0 to 10, with higher scores indicating better performance). Both metrics are assessed by human raters with four raters included in this study. Emotion response accuracy is first calculated within each dialogue and then averaged across dialogues for each rater. The final scores, averaged across all raters, will be reported.

\subsection{Baselines}
On the MELD dataset, 
we evaluate the proposed VL2E model by comparing it to the following models for ERC: DialogueRNN~\cite{majumder2019dialoguernn} models the speaker identity, conversation context and the corresponding emotion with RNN. 
ConGCN~\cite{zhang2019modeling} is a Graph Convolutional Network (GCN)-based model used to represent utterances, speakers and their relationships. MMGCN~\cite{hu2021mmgcn} is a model based on multimodal fused GCN that can effectively utilize both multimodal and long-distance contextual information. 
DAG-ERC~\cite{shen2021directed} is a directed acyclic neural network designed to better
model the intrinsic structure within a conversation. 
MM-DFN~\cite{hu2022mm} introduces a graph-based dynamic fusion module to fuse multimodal contextual features. M2FNet~\cite{chudasama2022m2fnet} employs a multi-head attention-based fusion mechanism to enhance the integration of multimodal features. 
EmoCaps~\cite{li2022emocaps} extracts multimodal emotion vectors using a structure named Emoformer. 
UniMSE~\cite{hu2022unimse} unifies multimodal sentiment analysis and ERC task through a T5-based framework~\cite{li2023ga2mif}. 
GA2MIF~\cite{li2023ga2mif} is a two-stage multi-source information fusion approach based on graph and attention.
FacialMMT~\cite{zheng2023facial} is a two-stage multimodal multi-task frame work that focuses on extracting faces sequences of active speakers and leverage frame-level facial expression recogniton tasks to help utterance-level emotion recognition. TelME~\cite{yun2024telme} incorporates cross-modal knowledge distillation to optimize the efficacy of weak modalities.

In the embodied system, UGotMe, we compare VL2E against TelME\textsuperscript{*}, a reproduced version without audio component to verify the effectiveness of our proposed denoising strategies.




%

\section{Results}

In this section, we aim to answer the following questions:
1) How well can VL2E perform on the dataset compared with other methods proposed for emotion recognition in conversation (ERC)? \\
2) How useful are the denoising strategies for both ERC on datasets and real-world human-robot conversations?\\
3) How well will UGotMe work in real-world affective human-robot interaction?


\begin{table*}[]
\caption{Comparison results for emotion recognition in conversation on MELD.}
    \centering
    \begin{tabular}{cccccccccc}
    \hline
     Models  & Modality &Neutral &Surprise &Fear &Sadness &Joy &Disgust &Anger  &F1  \\
     \hline
    DialogueRNN~\cite{majumder2019dialoguernn} &T+V+A &73.50 & 49.40 &1.20 &23.80 &50.70 &1.70 &41.50 &57.03 \\
     ConGCN~\cite{zhang2019modeling} &T+A &76.70 &50.30 &8.70 &28.50 &53.10 &10.60 &46.80 &59.40 \\
     MMGCN~\cite{hu2021mmgcn} &T+V+A &- &- &- &- &- &- &- &58.65 \\
     DAG-ERC~\cite{shen2021directed} &T&- &- &- &- &- &- &- &63.65 \\
    MM-DFN~\cite{hu2022mm} &T+V+A &77.76 &50.69 &- &22.94 &54.78 &- &47.82 &59.46 \\
    M2FNet~\cite{chudasama2022m2fnet} &T+V+A  &- &- &- &- &- &- &- &66.71 \\
    EmoCaps~\cite{li2022emocaps} & T+V+A &77.12 &\textbf{63.19} &3.03 &42.52 &57.50 &7.69 &57.54 &64.00 \\
    UniMSE~\cite{hu2022unimse} &T+V+A &- &- &- &- &- &- &- &65.51 \\
    GA2MIF~\cite{li2023ga2mif} & T+V+A &76.92 &49.08 &- &27.18 &51.87 &- &48.52 &58.94 \\
    FacialMMT~\cite{zheng2023facial} & T+V+A &80.13 &59.63 &19.18 &41.99 &64.88 &18.18 &\textbf{56.00} &66.58 \\
    TelME\textsuperscript{*}~\cite{yun2024telme} & T+V &79.85 &59.61 &\textbf{31.17} &{43.35} &{65.29} &{23.53} &55.00 &66.83\\
    \hline
    \textbf{VL2E} & T+V &\textbf{80.31} &{61.03} &14.71 &\textbf{46.28} &\textbf{65.33} &\textbf{28.57} &{55.07} &\textbf{67.29}\\
    \hline
    \end{tabular}
    \label{tab:compres}
\end{table*}

\begin{figure*}[t]
\centering
    \begin{subfigure}[b]{0.48\textwidth}
    \includegraphics[width=0.98\linewidth]{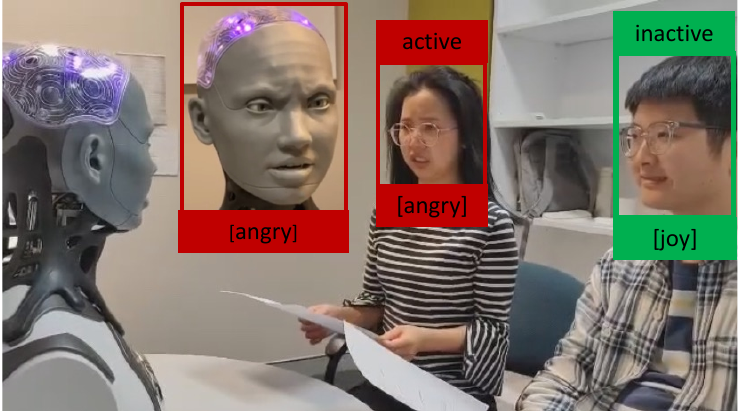}
    \caption{Ameca responds with an angry face when the active speaker said ``Dylon cheated me again, I don't wanna talk to him anymore" in anger.
     }
     \label{fig:comp-deploy1}
    \end{subfigure}
    ~
    \begin{subfigure}[b]{0.48\textwidth}
\includegraphics[width=0.98\linewidth]{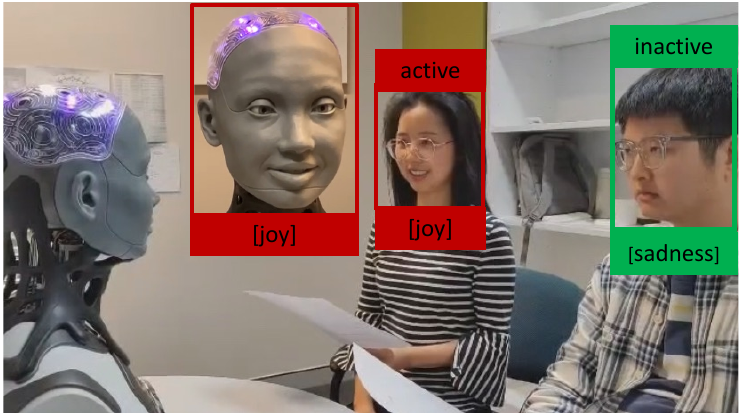}
    \caption{ Ameca responds with a joyful face when the active speaker said ``My paper was accepted finally" in joy.\\
    ~~~~~~
     }
     \label{fig:comp-deploy2}
    \end{subfigure}
\caption{Two examples of real-world execution of UGotMe. Active speakers who are conversing with Ameca are indicated by the red bounding boxes. }
\label{fig:deploy}
\end{figure*}

\begin{figure*}[h]
    \centering
    \begin{subfigure}[b]{0.48\textwidth}
    \includegraphics[width=0.98\linewidth]{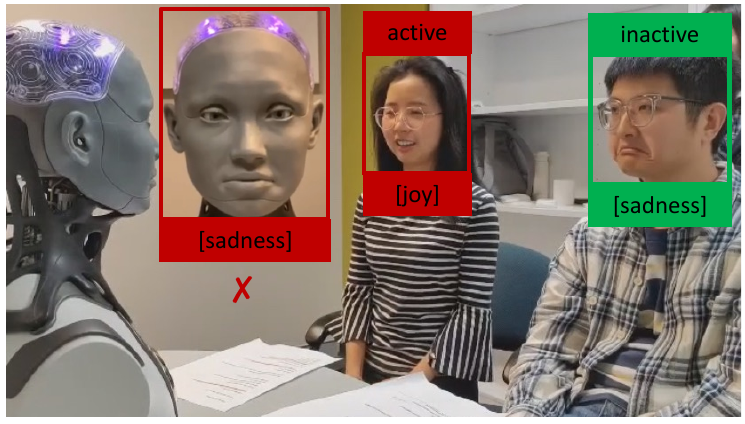}
    \caption{UGotMe-TelME.}
     \label{fig:comp-telme}
    \end{subfigure}
    ~
    \begin{subfigure}[b]{0.48\textwidth}
\includegraphics[width=0.98\linewidth]{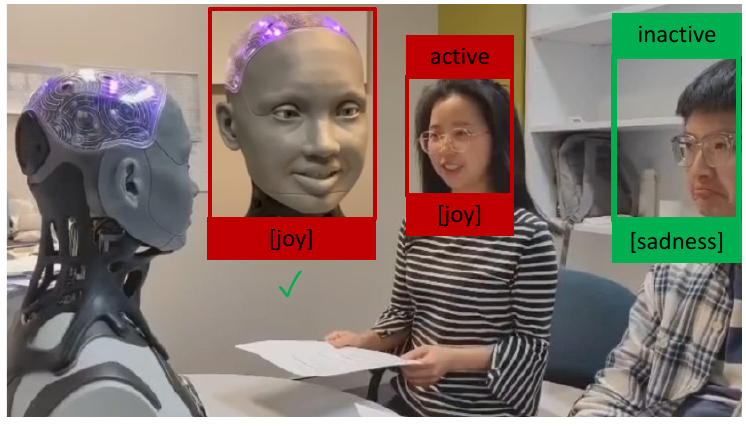}
    \caption{ UGotMe-VL2E.
     }
     \label{fig:comp-ours}
    \end{subfigure}
\caption{A comparison between UGotMe-TelME and UGotMe-VL2E. In both cases, the inactive speaker has a sad expression, while the active speaker, who is talking with Ameca, has a joyful expression. 
Dialogue context for both cases are: ``The movie we saw last night is really impressive.
That’s awesome. What movie did you watch? You jump, I jump". Ameca is supposed to deliver the same emotion as the active speaker through facial expression. However, in the case of UGotMe-TelME, distracting face confuses the model, leading to the wrong answer.}
\label{fig:comp-deploy}
\end{figure*}

\begin{table}[]
\caption{Comparison results in terms of emotion response accuracy and user experience score.}
    \centering
    \begin{tabular}{cccc}
    \hline
     Systems  & Accuracy $\uparrow$ & User experience$\uparrow$  \\
     \hline
    UGotMe-TelME & 50.63 & 6.15 \\
    UGotMe-VL2E\textsuperscript{1} & 59.08 & 6.63 \\
    \hline
    \textbf{UGotMe-VL2E} & \textbf{77.29} & \textbf{7.89} \\
    \hline
    \end{tabular}
    \label{tab:deploy-compres}
\end{table}

\begin{table}[]
\caption{Ablation study of VL2E based on F1 score.}
    \centering
    \begin{tabular}{cc}
    \hline
     \textbf{VL2E}  & \textbf{67.29}  \\ 
     \hline
    - w/o text &35.97 \\
    - w/o vision & 65.56 \\
    - w/o face extraction & 65.95 \\
    - w/o neutral normalization & 66.60 \\
    \hline
    \end{tabular}
    \label{tab:ablation}
\end{table}

\subsection{Benchmark results}
The performance of VL2E in comparison to other methods on the MELD dataset is summarized in Table~\ref{tab:compres}. The baselines tagged with \textsuperscript{*} denote reproduced versions, where we remove audio component to ensure consistency in the number of input modalities compared to VL2E. These baselines are also evaluated in real-world deployment scenarios. Other baselines represent the original models, and their results are obtained from referenced papers.
VL2E outperforms all baselines on MELD even though most of them employ 3 modalities in the comparison experiments.
As can be seen, there is an increase of 
$+10.26\%$ F1 compared to DialogueRNN; $+7.89\%$ F1 compared to ConGCN;
$+8.64\%$ F1 compared to MMGCN;
$+8.35\%$ F1 compared to GA2MIF; 
and $+0.71\%$ F1 compared to FacialMMT.
We attribute the higher performance of VL2E to the effective utilization of visual emotional cues through face extraction and person-specific neutral normalization. 


\subsection{Ablations}
We conduct an ablation study to inspect the impact of each modality, and verify the effectiveness of face extraction and person-specific neutral normalization. The results are summarized in Table~\ref{tab:ablation}. As can be observed, removing either the text or vision modality will impair performance. 
There is a decrease of $1.34\%$ in F1 score when face extraction is not applied, and visual features are extracted from the entire video frame instead. This highlights the significance of focusing on emotion-rich facial expressions. Additionally, the findings underscore the need for person-specific neutral normalization.


\subsection{Real-world Deployment}

We deploy UGotMe on a humanoid robot named Ameca to demonstrate its practical usage in the real world. 
To improve overall user experience and ensure fair comparison, we use GPT-4o as an auxiliary model for vision-only emotion recognition through all comparison experiments.
Two examples of real-world execution are shown in Fig.~\ref{fig:deploy}.
We also substitute the emotion recognition model in UGotMe, VL2E, with TelME and compare it against the original version. We use UGotMe-VL2E and UGotMe-TelME to distinguish between the two system versions, with UGotMe referring to UGotMe-VL2E unless otherwise specified.
Comparison results in terms of emotion response accuracy and user experience are summarized in Table~\ref{tab:deploy-compres}. 
We do not apply the customized active face extraction in UGotMe-VL2E\textsuperscript{1},  using it as one of our baselines to validate the effectiveness of the customized face extraction strategy. As can be seen, UGotMe-VL2E outperforms all baselines in both accuracy and user experience score with an increase of $26.66\%$ in accuracy and 1.74 in user experience score compared to UGotMe-TelME,
an increase of $18.21\%$ in accuracy and 1.26 in user experience compared to UGotMe-VL2E\textsuperscript{1}.
These results validate the effectiveness of our denoising strategies and demonstrate the feasibility of deploying the VL2E model in an embodied system.

Fig.~\ref{fig:comp-deploy} shows two cases where UGotMe-TelME fails
while UGotME-VL2E successes. In both cases, the inactive speaker (indicated by the green bounding box) displays a different emotion compared to the active speaker (indicated by the red bounding box). Distraction from facial expressions can significantly impact emotion recognition when no denoising strategies are applied, potentially leading to incorrect conclusions.

Models for emotion recognition are hosted on a local server with one NVIDIA RTX 4090 GPU.
Videos can be viewed at \href{https://lipzh5.github.io/HumanoidVLE/}{https://lipzh5.github.io/HumanoidVLE/}.

\section{Conclusion}
This paper presents UGotMe, an embodied system for affective human-robot interaction. Our key novelty is addressing embodiment issues for multimodal emotion recognition models
by designing denoising strategies customized for humanoid robot (Ameca, in the present study) in multiparty conversation scenarios. We also propose a multimodal emotion recognition model namded VL2E, designed to be compatible with denoising strategies. VL2E can be integrated into UGotMe for practical applications and demonstrates superiority over all baselines on the MELD dataset.
Real-world deployment experiments demonstrate that UGotMe effectively provides appropriate emotional responses to human interactants while maintaining a positive user experience, even in the presence of distracting factors.

UGotMe has a number of limitations that could be addressed by future work. First, robotic facial expressions are used to deliver emotions and generated in line with parallel empathy, i.e., generating the same emotion as the peer. Future work could explore new models to enable humanoid robots generating emotions in response to peer's emotion (in line with reactive empathy~\cite{davis2018empathy}). Furthermore, acoustic signals are not utilized and incorporated into the model to help understand human emotions. Future work will investigate the possibility of transmitting acoustic signals from our humanoid robot, Ameca, and further incorporating them into our future reactive emotion generation model.

\section{ACKNOWLEDGMENT}
This work was partly supported by Australian Research Council Discovery Grant DP240102050, LIEF Grant LE240100131 and Linkage Grant LP230201022. 









\bibliographystyle{IEEEtran}
\bibliography{main}

\begin{thebibliography}{10}
\providecommand{\url}[1]{#1}
\csname url@rmstyle\endcsname
\providecommand{\newblock}{\relax}
\providecommand{\bibinfo}[2]{#2}
\providecommand\BIBentrySTDinterwordspacing{\spaceskip=0pt\relax}
\providecommand\BIBentryALTinterwordstretchfactor{4}
\providecommand\BIBentryALTinterwordspacing{\spaceskip=\fontdimen2\font plus
\BIBentryALTinterwordstretchfactor\fontdimen3\font minus \fontdimen4\font\relax}
\providecommand\BIBforeignlanguage[2]{{%
\expandafter\ifx\csname l@#1\endcsname\relax
\typeout{** WARNING: IEEEtran.bst: No hyphenation pattern has been}%
\typeout{** loaded for the language `#1'. Using the pattern for}%
\typeout{** the default language instead.}%
\else
\language=\csname l@#1\endcsname
\fi
#2}}

\bibitem{cao24humanoid}
L.~Cao, ``{AI} robots and humanoid {AI:} review, perspectives and directions,'' \emph{CoRR}, vol. abs/2405.15775, 2024.

\bibitem{baltruvsaitis2015cross}
T.~Baltru{\v{s}}aitis, M.~Mahmoud, and P.~Robinson, ``Cross-dataset learning and person-specific normalisation for automatic action unit detection,'' in \emph{IEEE international conference and workshops on automatic face and gesture recognition (FG)}, 2015.

\bibitem{mehrabian2017communication}
A.~Mehrabian, ``Communication without words,'' in \emph{Communication theory}, 2017.

\bibitem{davis2018empathy}
M.~H. Davis, \emph{Empathy: A social psychological approach}.\hskip 1em plus 0.5em minus 0.4em\relax Routledge, 2018.

\bibitem{poria2018meld}
S.~Poria, D.~Hazarika, N.~Majumder, G.~Naik, E.~Cambria, and R.~Mihalcea, ``Meld: A multimodal multi-party dataset for emotion recognition in conversations,'' \emph{arXiv preprint arXiv:1810.02508}, 2018.

\bibitem{zhu2022affective}
H.~Zhu, C.~Yu, and A.~Cangelosi, ``Affective human-robot interaction with multimodal explanations,'' in \emph{International Conference on Social Robotics}, 2022.

\bibitem{chowdary2023deep}
M.~K. Chowdary, T.~N. Nguyen, and D.~J. Hemanth, ``Deep learning-based facial emotion recognition for human--computer interaction applications,'' \emph{Neural Computing and Applications}, 2023.

\bibitem{maeda2020human}
Y.~Maeda, T.~Sakai, K.~Kamei, and E.~W. Cooper, ``Human-robot interaction based on facial expression recognition using deep learning,'' in \emph{SCIS-ISIS}, 2020.

\bibitem{maeda2018human}
Y.~Maeda and S.~Geshi, ``Human-robot interaction using markovian emotional model based on facial recognition,'' in \emph{SCIS-ISIS}, 2018.

\bibitem{esfandbod2019human}
A.~Esfandbod, Z.~Rokhi, A.~Taheri, M.~Alemi, and A.~Meghdari, ``Human-robot interaction based on facial expression imitation,'' in \emph{ICRoM}, 2019.

\bibitem{ghorbandaei2018human}
A.~Ghorbandaei~Pour, A.~Taheri, M.~Alemi, and A.~Meghdari, ``Human--robot facial expression reciprocal interaction platform: case studies on children with autism,'' \emph{International Journal of Social Robotics}, 2018.

\bibitem{liu2017facial}
Z.~Liu, M.~Wu, W.~Cao, L.~Chen, J.~Xu, R.~Zhang, M.~Zhou, and J.~Mao, ``A facial expression emotion recognition based human-robot interaction system.'' \emph{IEEE CAA J. Autom. Sinica}, 2017.

\bibitem{chen2022k}
L.~Chen, K.~Wang, M.~Li, M.~Wu, W.~Pedrycz, and K.~Hirota, ``K-means clustering-based kernel canonical correlation analysis for multimodal emotion recognition in human--robot interaction,'' \emph{IEEE Transactions on Industrial Electronics}, 2022.

\bibitem{cid2015novel}
F.~Cid, L.~J. Manso, and P.~N{\'u}nez, ``A novel multimodal emotion recognition approach for affective human robot interaction,'' \emph{Proceedings of fine}, 2015.

\bibitem{jung2004affective}
H.-W. Jung, Y.-H. Seo, M.~S. Ryoo, and H.~S. Yang, ``Affective communication system with multimodality for a humanoid robot, ami,'' in \emph{IEEE/RAS International Conference on Humanoid Robots}, 2004.

\bibitem{ma2020survey}
Y.~Ma, K.~L. Nguyen, F.~Z. Xing, and E.~Cambria, ``A survey on empathetic dialogue systems,'' \emph{Information Fusion}, 2020.

\bibitem{yun2024telme}
T.~Yun, H.~Lim, J.~Lee, and M.~Song, ``Telme: Teacher-leading multimodal fusion network for emotion recognition in conversation,'' \emph{arXiv preprint arXiv:2401.12987}, 2024.

\bibitem{hu2022unimse}
G.~Hu, T.-E. Lin, Y.~Zhao, G.~Lu, Y.~Wu, and Y.~Li, ``Unimse: Towards unified multimodal sentiment analysis and emotion recognition,'' \emph{arXiv preprint arXiv:2211.11256}, 2022.

\bibitem{li2022emocaps}
Z.~Li, F.~Tang, M.~Zhao, and Y.~Zhu, ``Emocaps: Emotion capsule based model for conversational emotion recognition,'' \emph{arXiv preprint arXiv:2203.13504}, 2022.

\bibitem{zheng2023facial}
W.~Zheng, J.~Yu, R.~Xia, and S.~Wang, ``A facial expression-aware multimodal multi-task learning framework for emotion recognition in multi-party conversations,'' in \emph{ACL}, 2023.

\bibitem{chudasama2022m2fnet}
V.~Chudasama, P.~Kar, A.~Gudmalwar, N.~Shah, P.~Wasnik, and N.~Onoe, ``M2fnet: Multi-modal fusion network for emotion recognition in conversation,'' in \emph{CVPR}, 2022.

\bibitem{zhang2016joint}
K.~Zhang, Z.~Zhang, Z.~Li, and Y.~Qiao, ``Joint face detection and alignment using multitask cascaded convolutional networks,'' \emph{IEEE signal processing letters}, 2016.

\bibitem{baltruvsaitis2016openface}
T.~Baltru{\v{s}}aitis, P.~Robinson, and L.-P. Morency, ``Openface: an open source facial behavior analysis toolkit,'' in \emph{IEEE winter conference on applications of computer vision}, 2016.

\bibitem{5543148}
S.~Zafeiriou and M.~Petrou, ``Sparse representations for facial expressions recognition via l1 optimization,'' in \emph{CVPR}, 2010.

\bibitem{szegedy2017inception}
C.~Szegedy, S.~Ioffe, V.~Vanhoucke, and A.~Alemi, ``Inception-v4, inception-resnet and the impact of residual connections on learning,'' in \emph{AAAI}, 2017.

\bibitem{yi2014learning}
D.~Yi, Z.~Lei, S.~Liao, and S.~Z. Li, ``Learning face representation from scratch,'' \emph{arXiv preprint arXiv:1411.7923}, 2014.

\bibitem{song2022supervised}
X.~Song, L.~Huang, H.~Xue, and S.~Hu, ``Supervised prototypical contrastive learning for emotion recognition in conversation,'' \emph{arXiv preprint arXiv:2210.08713}, 2022.

\bibitem{liu2023pre}
P.~Liu, W.~Yuan, J.~Fu, Z.~Jiang, H.~Hayashi, and G.~Neubig, ``Pre-train, prompt, and predict: A systematic survey of prompting methods in natural language processing,'' \emph{ACM Computing Surveys}, 2023.

\bibitem{gao2021simcse}
T.~Gao, X.~Yao, and D.~Chen, ``Simcse: Simple contrastive learning of sentence embeddings,'' \emph{arXiv preprint arXiv:2104.08821}, 2021.

\bibitem{tsai2019multimodal}
Y.-H.~H. Tsai, S.~Bai, P.~P. Liang, J.~Z. Kolter, L.-P. Morency, and R.~Salakhutdinov, ``Multimodal transformer for unaligned multimodal language sequences,'' in \emph{ACL}, 2019.

\bibitem{majumder2019dialoguernn}
N.~Majumder, S.~Poria, D.~Hazarika, R.~Mihalcea, A.~Gelbukh, and E.~Cambria, ``Dialoguernn: An attentive rnn for emotion detection in conversations,'' in \emph{AAAI}, 2019.

\bibitem{zhang2019modeling}
D.~Zhang, L.~Wu, C.~Sun, S.~Li, Q.~Zhu, and G.~Zhou, ``Modeling both context-and speaker-sensitive dependence for emotion detection in multi-speaker conversations,'' in \emph{IJCAI}, 2019.

\bibitem{hu2021mmgcn}
J.~Hu, Y.~Liu, J.~Zhao, and Q.~Jin, ``Mmgcn: Multimodal fusion via deep graph convolution network for emotion recognition in conversation,'' \emph{arXiv preprint arXiv:2107.06779}, 2021.

\bibitem{shen2021directed}
W.~Shen, S.~Wu, Y.~Yang, and X.~Quan, ``Directed acyclic graph network for conversational emotion recognition,'' \emph{arXiv preprint arXiv:2105.12907}, 2021.

\bibitem{hu2022mm}
D.~Hu, X.~Hou, L.~Wei, L.~Jiang, and Y.~Mo, ``Mm-dfn: Multimodal dynamic fusion network for emotion recognition in conversations,'' in \emph{ICASSP}, 2022.

\bibitem{li2023ga2mif}
J.~Li, X.~Wang, G.~Lv, and Z.~Zeng, ``Ga2mif: graph and attention based two-stage multi-source information fusion for conversational emotion detection,'' \emph{IEEE Transactions on affective computing}, 2023.

\end{thebibliography}

\end{document}